\documentclass[conference]{IEEEtran}
\IEEEoverridecommandlockouts

\usepackage{cite}
\usepackage{amsmath,amssymb,amsfonts}
\usepackage{graphicx}
\usepackage{textcomp}
\usepackage{xcolor}
\usepackage{booktabs}
\usepackage{tabularx}
\usepackage{array}
\usepackage{url}

\newcolumntype{Y}{>{\raggedright\arraybackslash}X}
\def\BibTeX{{\rm B\kern-.05em{\sc i\kern-.025em b}\kern-.08em
    T\kern-.1667em\lower.7ex\hbox{E}\kern-.125emX}}

\begin{document}

\title{
GraspTune: Tactile-Driven Execution\\
Refinement for Robust Grasping
}

\author{
\IEEEauthorblockN{
Juntao Li\textsuperscript{1},
Xingke Xia\textsuperscript{1},
Sichao Liu\textsuperscript{2},
Daqiang Guo\textsuperscript{1,*}
}

\IEEEauthorblockA{
\textsuperscript{1}
The Hong Kong University of Science and Technology (Guangzhou), Guangzhou, China\\
\textsuperscript{2}
KTH Royal Institute of Technology, Stockholm, Sweden
}

\thanks{\textsuperscript{*} Corresponding author: Daqiang Guo
(e-mail: daqiangguo@hkust-gz.edu.cn).}
}

\maketitle

\begin{abstract}
Visual grasp proposal generation has advanced rapidly, yet converting a selected proposal into a stable physical grasp remains a central execution-stage challenge. This paper introduces GraspTune, a tactile-driven execution-stage refinement framework that starts from a nominal proposal and applies bounded residual TCP motions during approach, contact formation, and final grasp execution. GraspTune learns control-facing contact semantics from local depth, tactile signals, state, and history using state-conditioned expert contact queries and multi-task supervision for contact change, contact risk, and post-close readiness. The representation conditions a diffusion-pretrained residual policy and is aligned with PPO for closed-loop execution. Across more than 60,000 simulated executions over 20 object categories, GraspTune establishes an execution-layer benefit across four proposal generators, raising stable grasp success by \(+19.22\), \(+9.55\), \(+12.45\), and \(+20.70\) percentage points for GraspNet, Contact-GraspNet, AnyGrasp, and VGN. A four-fold held-out category study raises unseen-object execution from \(54.58\%\) to \(70.33\%\), showing category-disjoint generalization of contact correction. Across more than 1,000 real-robot trials on a UR5e setup with Xense fingertip sensors, GraspTune raises GraspNet execution from \(71.0\%\) to \(84.3\%\), validating direct transfer without real-world policy fine-tuning. Together, these results turn visually plausible proposals into stable physical grasps for downstream contact-rich manipulation. A supplementary video is available at
\url{https://youtu.be/kcq7fSLNtzU}.
\end{abstract}

\begin{IEEEkeywords}
robotic grasping, tactile sensing, tactile-driven policy, contact-rich manipulation, grasp refinement
\end{IEEEkeywords}

\section{Introduction}

Robotic grasping is the first physical commitment in many manipulation pipelines. Downstream contact plans start from the grasp they receive, so weak, one-sided, or object-displacing contact can propagate through the rest of the task.

Visual grasp proposal methods have made this step much stronger. GraspNet~\cite{fang2020graspnet}, AnyGrasp~\cite{fang2023anygrasp}, Contact-GraspNet~\cite{sundermeyer2021contactgraspnet}, VGN~\cite{breyer2021vgn}, GIGA~\cite{jiang2021giga}, and GraspNeRF~\cite{dai2023graspnerf} produce 6-DoF grasps from point clouds, RGB-D, volumetric fields, implicit geometry, or multi-view reconstruction. Their output is a strong geometric starting point, but execution must still handle close-range occlusion, first fingertip contact, object motion, closure, and lift.

\begin{figure}[t]
    \centering
    \includegraphics[width=0.78\columnwidth]{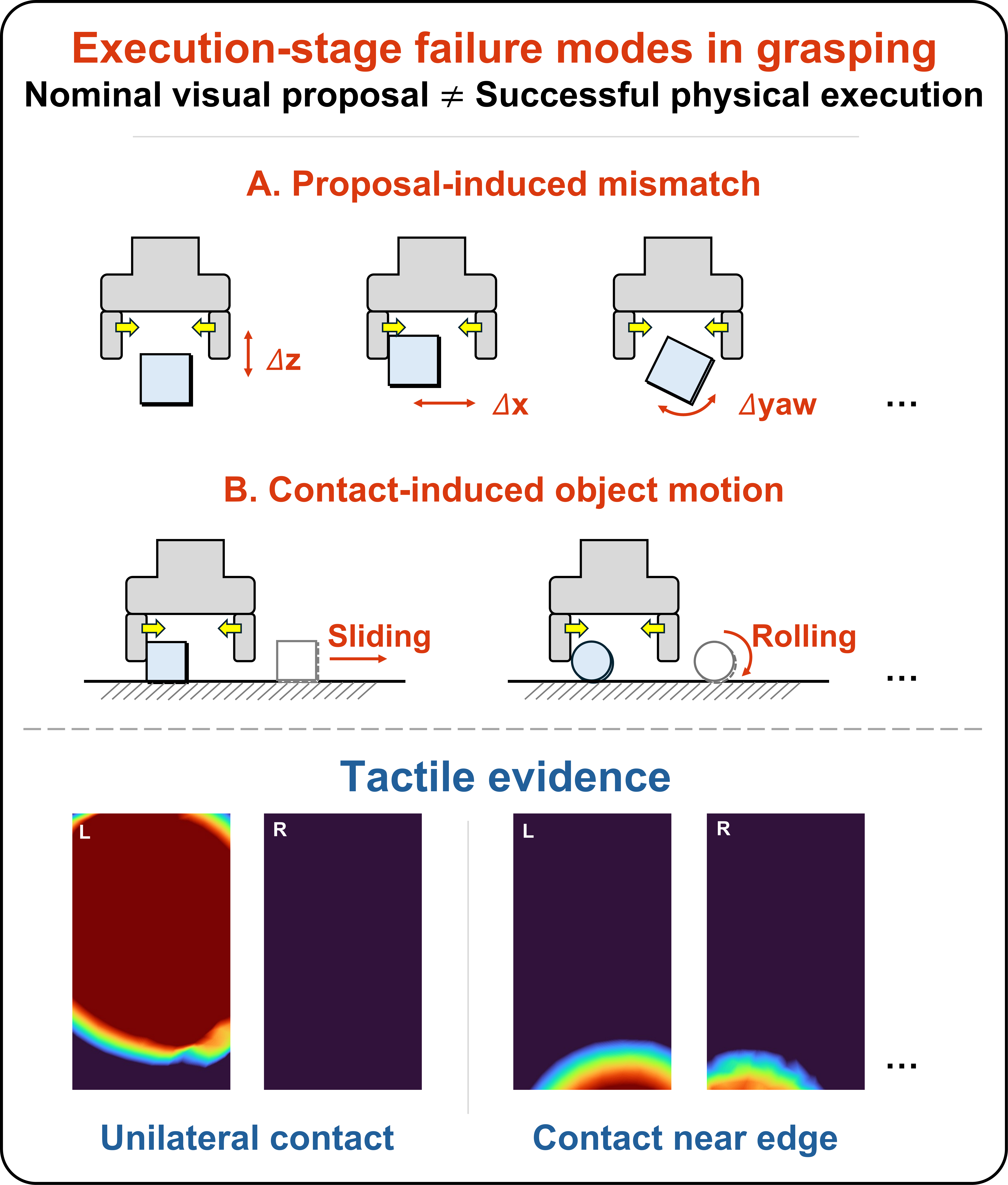}
    \caption{Execution-stage failure modes in grasping.}
    \label{fig:motivation}
\end{figure}

This paper targets that physical execution stage. After first touch, the gripper occludes the object, depth becomes less informative, and the object may slide, roll, or rotate. Fig.~\ref{fig:motivation} shows how a proposal can remain geometrically plausible while tactile evidence reveals unilateral contact, edge-biased pressure, or an unstable grasp state. The key question becomes: what contact has formed, and what residual motion will stabilize it?

We study \emph{tactile-driven execution-stage grasp refinement}: starting from a nominal visual proposal, the robot adjusts bounded residual TCP motions using local depth, bilateral tactile signals, execution state, and short history. GraspTune converts visually plausible 6-DoF proposals into stable physical grasps across proposal generators, occlusion levels, unseen objects, and a real-robot platform.

The method builds contact representations for control. Contact-Semantic Representation uses state-conditioned expert contact queries to read tactile streams and visual context, then predicts action-conditioned contact consequences under short residual action chunks. Residual Policy Pretraining learns a diffusion residual controller from demonstrations, and Policy Adaptation turns it into a PPO closed-loop execution policy~\cite{chi2023diffusion,schulman2017ppo}.

The paper makes three contributions:
\begin{itemize}
    \item We formulate tactile-driven execution-stage refinement for nominal 6-DoF grasp proposals, targeting the physical contact state that determines stable grasping.
    \item We introduce GraspTune, which combines Contact-Semantic Representation, Residual Policy Pretraining, and Policy Adaptation for tactile-driven residual control.
    \item We establish effectiveness and generality at scale: more than 60,000 simulated executions improve stable grasp success by \(+19.22\), \(+9.55\), \(+12.45\), and \(+20.70\) percentage points after GraspNet, Contact-GraspNet, AnyGrasp, and VGN, respectively; a four-fold category-disjoint study raises unseen-object execution from \(54.58\%\) to \(70.33\%\); and more than 1,000 real-robot trials raise GraspNet execution from \(71.0\%\) to \(84.3\%\) without real-world policy fine-tuning.
\end{itemize}

\section{Related Work}

\subsection{Grasp Proposal Generation}

Grasp proposal generators answer a pre-contact geometric question: where can the gripper close? Earlier systems combined geometric candidates, analytic quality metrics, and learned scoring~\cite{tenpas2017gpd,mahler2017dexnet,liang2019pointnetgpd,mahler2019ambidextrous}. Modern 6-DoF pipelines predict or rank grasps from point clouds, RGB-D, volumetric fields, implicit geometry, or multi-view reconstruction~\cite{fang2020graspnet,mousavian20196dofgraspnet,sundermeyer2021contactgraspnet,fang2023anygrasp,qin2020s4g,breyer2021vgn,jiang2021giga,dai2023graspnerf}. Their geometric scores precede fingertip contact and contact-induced object motion. Converting an admitted 6-DoF proposal into a stable physical hold therefore remains an execution-stage problem.

\subsection{Closed-Loop Grasping}

Closed-loop grasping updates the robot during execution through visuomotor control~\cite{levine2018handeye,kalashnikov2018qtopt}, depth-based servoing~\cite{morrison2018ggcnn,haviland2020finalphase}, tactile servoing and MPC~\cite{li2013tactileservoing,tian2019manipulation}, contact-state regulation~\cite{hogan2020tactile}, or visuo-tactile regrasping~\cite{calandra2018more}. Recent systems combine semantic perception with adaptive force control and support diverse or deformable-object grasping~\cite{li2026goto,wang2025d3grasp}; TacRefineNet studies tactile-only, goal-conditioned local alignment for edge-prominent objects~\cite{wang2025tacrefinenet}. These approaches establish the value of execution feedback, while tactile formation of a stable hold from an externally admitted proposal remains a distinct open problem.

\subsection{Visuo-Tactile Policies for Contact-Rich Manipulation}

Recent visuo-tactile policies use touch for broader contact-rich manipulation: multimodal representation learning~\cite{lee2019making}, inference-time visuomotor steering~\cite{zhang2026touchguide}, slow-fast diffusion control~\cite{xue2025rdp}, tactile residual adaptation of visual policies~\cite{yu2026omnitactune}, or tactile feedback for VLA policies~\cite{ma2026taccorl}. These policies optimize task-action streams once a workable grasp has been established. In practice, unstable contact formation can undermine that precondition before downstream manipulation begins. GraspTune addresses this missing execution layer with expert contact queries and action-conditioned contact-consequence supervision, which diffusion and PPO convert into a closed-loop residual controller~\cite{chi2023diffusion,schulman2017ppo}.

\section{Methodology}

\begin{figure*}[t]
    \centering
    \includegraphics[width=\textwidth]{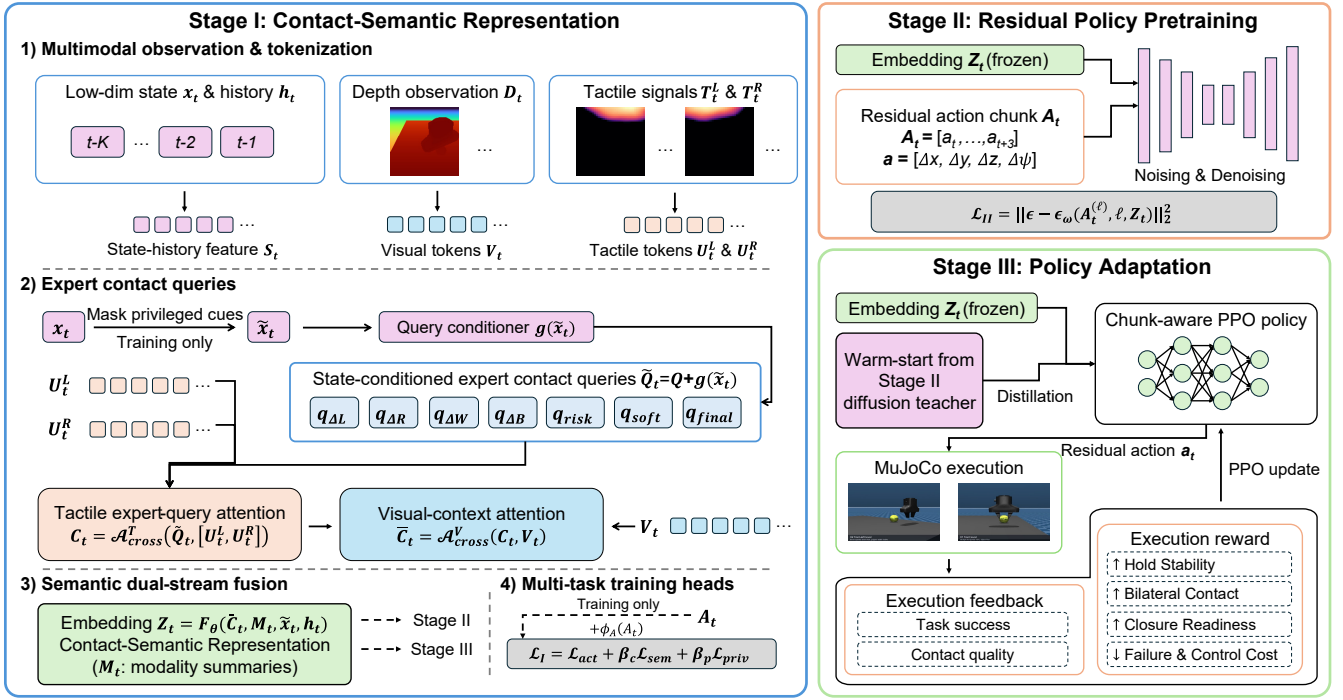}
    \caption{GraspTune: Tactile-driven execution-stage refinement framework.}
    \label{fig:framework}
\end{figure*}

\subsection{Problem Formulation}
\label{sec:problem_formulation}

We formulate grasp execution as tactile-driven refinement. A visual module provides a nominal grasp, whose execution can still fail when occlusion, object geometry, surface material, pose error, or contact-induced motion produces unilateral, migrating, or poorly balanced finger contact. The refinement policy uses local tactile feedback to convert the proposal into a contact state that supports stable grasping.

Let \(g_0\) denote the nominal 6-DoF grasp from the visual proposal generator. Starting from \(g_0\), GraspTune applies local residual TCP motions that define the executed grasp \(g_{\mathrm{exec}}\). At refinement step \(t\), the policy input is
\begin{equation}
    o_t = \{D_t,T_t^L,T_t^R,\tilde{x}_t,h_t\}.
\end{equation}
Here \(D_t\) is a local depth image, \(T_t^L,T_t^R\) are left and right tactile maps, \(\tilde{x}_t\) contains bilateral contact area, centroid, peak response, balance, TCP residual pose, finger state, and trigger flags, and \(h_t\) stores short tactile, contact, TCP, and finger history. Explicit maps preserve fingertip contact geometry, while state and history summarize execution evidence for stable grasping.

During data collection, the full low-dimensional record can be written as \(x_t=[\tilde{x}_t,p_t]\), where \(p_t\) is an eight-dimensional privileged cue produced by the scripted controller:
\[
p_t=[u_x,u_y,u_z,u_\psi,\Delta c,s_f,I_{\mathrm{freeze}},I_{\mathrm{strong}}].
\]
Here \(u_x,u_y,u_z,u_\psi\) are normalized scripted residual actions, \(\Delta c\) is the closing-increment cue, \(s_f\) is the forward-scale cue, and the last two entries indicate unilateral-approach freezing and strong unilateral contact. The encoder receives \(\tilde{x}_t\); an auxiliary head reconstructs \(p_t\), distilling scripted contact cues into \(Z_t\) while preserving the test-time policy input.

The policy output is a bounded local residual motion,
\begin{equation}
    a_t = [\Delta x,\Delta y,\Delta z,\Delta \psi],
\end{equation}
The translational terms act in the local TCP frame, and \(\Delta\psi\) rotates about the local \(z\)-axis. Per-step limits are \(\lvert\Delta x\rvert,\lvert\Delta y\rvert\leq1.5\) mm, \(0\leq\Delta z\leq0.4\) mm, and \(\lvert\Delta\psi\rvert\leq2^\circ\), with contact-dependent forward-motion clipping. The executed pose stays within \(\pm15\) mm laterally, \([0,5]\) mm axially, and \(\pm5^\circ\) of the proposal. In simulation, over a maximum horizon of \(N_{\mathrm{ref}}=200\) refinement steps, the objective maximizes final stable-hold success:
\begin{equation}
    \pi^\star =
    \arg\max_{\pi}
    \mathbb{E}_{\tau\sim\pi}
    \left[\mathbb{I}\left(\mathrm{hold}_{3s}(\tau)=1\right)\right].
\end{equation}
Vision provides the nominal grasp; closed-loop tactile refinement determines the final contact state.

\subsection{Framework}

Fig.~\ref{fig:framework} summarizes the proposed framework. Its central design choice is to learn \emph{control-facing contact semantics} instead of a generic multimodal embedding. Execution-stage grasp refinement depends on weak-side engagement, bilateral balance, contact migration, local indentation risk, and closure readiness. GraspTune keeps these variables available to the controller through structured contact queries and explicit modality summaries. We organize the framework around three stages: Contact-Semantic Representation, Residual Policy Pretraining, and Policy Adaptation.

\textbf{Stage I: Contact-Semantic Representation.}
Stage I maps \(o_t\) to a contact-semantic latent state \(Z_t=f_\theta(o_t)\). Depth, left tactile, right tactile, low-dimensional state, and history are encoded through separate pathways:
\begin{equation}
    V_t = E_D(D_t), \qquad
    U_t^L = E_T(T_t^L), \qquad
    U_t^R = E_T(T_t^R),
\end{equation}
where \(V_t\) denotes visual tokens and \(U_t^L,U_t^R\) denote left and right tactile tokens. The tactile encoder \(E_T\) is shared, and the two tactile streams remain named separately until they form the bilateral tactile context for the query block. The policy-visible state carries side information by storing left and right contact variables in distinct entries, and the ordered left and right summaries used in the final fusion. The query block therefore aggregates bilateral tactile evidence, while the state and modality-summary paths provide the explicit interface for comparing the two fingers.

Stage I uses a set of role-specific contact queries:
\begin{equation}
    Q =
    (q_{\Delta L},q_{\Delta R},q_{\Delta W},
    q_{\Delta B},q_{\mathrm{risk}},
    q_{\mathrm{soft}},q_{\mathrm{final}}).
\end{equation}
The seven slots supervise future left-contact, right-contact, weaker-side, and balance changes, local contact-risk response, soft closure readiness, and strict/post-close readiness.

The queries are conditioned on the current execution state before they read tactile evidence:
\begin{equation}
    \tilde{Q}_t = Q + g_x(\tilde{x}_t).
\end{equation}
State conditioning lets the queries adapt their tactile readout to the measured execution phase. We use residual cross-attention blocks with pre-normalization:
\begin{equation}
\begin{aligned}
R &= X + \operatorname{Attn}(\operatorname{LN}_q(X), \operatorname{LN}_c(Y)),\\
\mathcal A_{\mathrm{cross}}(X,Y) &= R + \operatorname{FFN}(\operatorname{LN}_f(R)).
\end{aligned}
\end{equation}
Here \(\operatorname{Attn}(Q,C)\) uses \(Q\) as queries and maps \(C\) to keys and values through learned projections; \(\operatorname{FFN}\) uses a fourfold hidden expansion. The queries first read bilateral tactile evidence and then retrieve visual context:
\begin{equation}
\begin{aligned}
C_t&=\mathcal A_{\mathrm{cross}}^T(\tilde Q_t,[U_t^L,U_t^R]),\\
\bar C_t&=\mathcal A_{\mathrm{cross}}^V(C_t,V_t).
\end{aligned}
\end{equation}
The two blocks have separate parameters. Their residual paths preserve tactile-grounded query features while adding depth context. Supervised slots organize left-contact, right-contact, weak-side, balance, risk, and readiness consequences; ordered state and summary paths carry side information into the final representation.

The semantic stream combines both query readouts and scales them by measured contact strength:
\begin{equation}
\begin{aligned}
\alpha_t&=1-\exp[-(S_t^L+S_t^R)/10],\\
s_t^C&=\alpha_t\operatorname{LN}\!\left(\frac{\langle C_t\rangle+\langle\bar C_t\rangle}{2}\right),
\end{aligned}
\end{equation}
where \(S_t^L,S_t^R\) are nonnegative contact areas in pixels and \(\langle\cdot\rangle\) denotes mean pooling over the seven query-token slots. The gate activates as contact develops. Raw modality summaries remain available throughout execution:
\begin{equation}
M_t=[s_t^D,s_t^L,s_t^R,s_t^T,|s_t^L-s_t^R|],
\end{equation}
where each summary averages its modality tokens and \(s_t^T\) averages the concatenated bilateral tokens. A joint projection \(P_\theta\), state-conditioned FiLM modulation, and history projection \(E_h\) produce the final representation:
\begin{equation}
Z_t=\operatorname{LN}\!\left[\operatorname{FiLM}(P_\theta[M_t,s_t^C],\tilde x_t)+E_h(h_t)\right].
\end{equation}
FiLM applies \(u\odot[1+0.1\tanh\gamma(\tilde x_t)]+0.1\beta(\tilde x_t)\), with learned modulation functions. Tokens have 128 dimensions and \(Z_t\) has 256 dimensions.

The contact-query head predicts short-term contact changes, peak contact risk, and terminal post-close readiness conditioned on a residual action chunk:
\begin{equation}
    A_t = [a_t,\ldots,a_{t+H-1}],
\end{equation}
where each \(a_t\) is a 4-DoF residual TCP action. The chunk is projected into query-token space and added to every contact slot; five regression slots and two readiness slots use separate heads:
\begin{equation}
\begin{aligned}
    \tilde C_t^i &= C_t^i+\phi_A(A_t), \quad i=1,\ldots,7,\\
    \hat r_t^i &= g_i(\tilde C_t^i), \quad i=1,\ldots,5,\\
    \hat b_t^{\mathrm{soft}} &= g_6(\tilde C_t^6),\qquad
    [\hat b_t^{\mathrm{strict}},\hat b_t^{\mathrm{post}}] = g_7(\tilde C_t^7).
\end{aligned}
\end{equation}
During supervised training, \(A_t\) contains the next \(H=4\) demonstrated residual actions. The \(\hat r_t^{1:5}\) targets are left-contact, right-contact, weaker-side-contact, balance, and local-risk changes. Area changes use \((S_{t'}^j-S_t^j)/64\), clipped to \([-4,4]\), for \(j\in\{L,R,W\}\), with \(S^W=\min(S^L,S^R)\) and \(t'=\min(t+H,N_\tau-1)\). Balance change is clipped to \([-1,1]\). The risk target is the maximum peak indentation across the two tactile sides over \([t,t']\), normalized by a fixed training scale and clipped to \([0,1]\). Binary targets encode soft readiness, strict readiness, and episode-level post-close readiness. Near termination, the chunk repeats the last available action and targets use the available window. At deployment, the residual policy acts on the consequence-aware embedding \(Z_t\).

Stage I also reconstructs the training-only privileged cue \(p_t\) from \(Z_t\), forcing the representation to encode scripted contact cues from deployable inputs. The Stage-I loss is
\begin{equation}
    \mathcal{L}_{\mathrm{I}} =
    \mathcal{L}_{\mathrm{act}}
    + \beta_c \mathcal{L}_{\mathrm{sem}}
    + \beta_p \mathcal{L}_{\mathrm{priv}},
\end{equation}
where \(\mathcal{L}_{\mathrm{act}}\) is a smooth-\(L_1\) loss on demonstrated residual actions, \(\mathcal{L}_{\mathrm{sem}}=\sum_{i=1}^{5}\mathrm{MSE}(\hat r_t^i,r_t^i)+\mathrm{BCE}([\hat b_t^{\mathrm{soft}},\hat b_t^{\mathrm{strict}},\hat b_t^{\mathrm{post}}],[b_t^{\mathrm{soft}},b_t^{\mathrm{strict}},b_t^{\mathrm{post}}])\), and \(\mathcal{L}_{\mathrm{priv}}\) is a smooth-\(L_1\) reconstruction loss for the masked privileged cue. Later stages use the contact-semantic embedding \(Z_t\).

\textbf{Stage II: Residual Policy Pretraining.}
Stage II trains a temporal residual policy from demonstrations while keeping the Stage-I encoder frozen. Category labels are excluded, so the policy learns from contact evolution, execution state, and history. Given \(Z_t\), the diffusion policy predicts
\begin{equation}
    A_t = [a_t,\ldots,a_{t+H-1}],
\end{equation}
where each element is a 4-DoF residual TCP action. A conditional 1-D diffusion model perturbs clean chunks and predicts the injected noise:
\begin{equation}
    \mathcal{L}_{\mathrm{II}} =
    \mathbb{E}\left[\left\|\epsilon -
    \epsilon_\omega(A_t^{(\ell)}, \ell, Z_t)\right\|_2^2\right],
\end{equation}
where \(\ell\) is the diffusion step and \(\epsilon_\omega\) is the denoising network. The pretrained diffusion policy provides a multi-step residual behavior model grounded in tactile execution state.

\textbf{Stage III: Policy Adaptation.}
Stage III distills the diffusion behavior model into a PPO-compatible residual controller and adapts it through simulated closed-loop interaction. The actor executes one residual action per step, while an auxiliary chunk head preserves the multi-step teacher during warm start. Let \(\hat{A}_t^{\mathrm{diff}}\) be the diffusion teacher chunk and \(\hat{a}_{t,0}^{\mathrm{diff}}\) its first action. The warm-start objective is
\begin{equation}
    \begin{aligned}
    \mathcal{L}_{\mathrm{distill}} &=
    \rho\!\left(\mu_\phi(Z_t),\hat{a}_{t,0}^{\mathrm{diff}}\right) \\
    &\quad + \lambda_c\rho\!\left(A_\phi(Z_t),\hat{A}_t^{\mathrm{diff}}\right) \\
    &\quad + \lambda_s\left\|A_\phi(Z_t)_0-\mu_\phi(Z_t)\right\|_2^2 ,
    \end{aligned}
\end{equation}
where \(\mu_\phi\) is the single-step PPO actor, \(A_\phi\) is the auxiliary chunk head, and \(\rho\) denotes the smooth-\(L_1\) action regression loss. After warm start, PPO updates the residual controller with an execution reward:
\begin{equation}
    \begin{aligned}
    r_t &= w_h I_{\mathrm{hold}}
    + w_b q_{\mathrm{bilateral}}
    + w_r q_{\mathrm{ready}}
    + w_p q_{\mathrm{post}} \\
    &\quad - w_d I_{\mathrm{drop}}
    - w_\tau I_{\mathrm{timeout}}
    - w_i q_{\mathrm{imbalance}}
    - w_a \|a_t\|_2^2 .
    \end{aligned}
\end{equation}
Here \(q_{\mathrm{bilateral}}\) rewards bilateral support through weak-side contact and bottleneck area, \(q_{\mathrm{ready}}\) rewards soft or strict closing readiness, and \(q_{\mathrm{post}}\) scores post-close contact quality using weaker-side area and balance. Deployment executes the PPO actor without scripted TCP-action guidance.

\subsection{Deployment}

The visual proposal generator supplies a reference grasp, followed by approach, tactile refinement, final closure, and lift. The PPO actor controls four-dimensional TCP residuals; a contact-dependent schedule controls finger closure. Closure advances from bilateral engagement, unilateral contact, and recent history. Strong unilateral loading can pause closure and forward motion while the policy adjusts the TCP; bilateral readiness triggers final closure. At the simulation refinement limit, the controller performs final closure, post-close assessment, and the evaluation lift. The proposal remains fixed throughout refinement.

\textbf{Contact criteria.}
In simulation, contact areas count pixels with indentation above 0.1 mm in the \(200\times350\) sensing maps before resizing. Balance is \(B_t=|S_t^L-S_t^R|/(S_t^L+S_t^R)\), set to zero at zero total area. Strict readiness requires both areas to reach 60 pixels with \(B_t\leq0.90\) for two consecutive frames. Soft readiness requires 80 stronger-side pixels, 20 weaker-side pixels, peak indentation of at least 0.5 mm, and either \(B_t\leq0.92\) or a 20-pixel increase in total or weaker-side area relative to the preceding history entry. Post-close readiness applies \(\min(S^L,S^R)\geq60\) pixels and \(B\leq0.72\). Positive bilateral area is required for readiness and post-close balance reward.

\textbf{Deployment interface.}
The policy receives one local depth view, two \(64\times64\) tactile maps, the policy-visible execution vector, and a three-step history. MuJoCo uses a 2 ms timestep with five physics steps per control action, giving a 100 Hz refinement loop. On hardware, 10 Hz denotes the complete action-update loop after tactile acquisition, preprocessing, policy inference, UR5e command generation, and gripper update. Each step uses the latest synchronized left--right Xense pair; we issue UR5e and gripper commands after observing and computing the policy action. Xense calibration and logging use a 0.05 s sensor sampling period. A normalized policy action \(a\in[-1,1]^4\) is converted to a local TCP increment by scaling its translational components by 2.5 mm and its yaw component by \(1^\circ\). A per-command safety limit bounds the translational and rotational increment norms to 3 mm and \(2^\circ\), respectively; the proposal-relative execution envelope is defined above. Commands use 0.012 m/s velocity and 0.015 m/s\(^2\) acceleration, with a calibrated sensor--TCP axis convention. All training and real-robot inference use an NVIDIA RTX 4060 Ti GPU with 16 GB memory. For sensor transfer, simulation passes FEM indentation maps in meters. On hardware, \(I_s\) and \(I_s^0\) are the raw three-channel Xense Difference image and no-load baseline for side \(s\). We compute \(r_s(u,v)=\operatorname{mean}_c|I_s(u,v,c)-I_s^0(u,v,c)|\), resize it to the policy grid, and map it as \(T_s=10^{-5}r_s\) m. The coefficient \(10^{-5}\) is the configured contact-strength proxy in meters per uint8 Difference unit, placing real Difference peaks on the indentation-equivalent scale used by the simulation-trained encoder. The peak map value enters the peak-response state feature, while adaptive baseline thresholds determine contact area. Raw execution closes from the nominal proposal; GraspTune closes from the refined state. Both use the same lift trajectory and 3 s stable-hold criterion.

\section{Experiments}

We evaluate GraspTune through four questions. \textbf{Q1:} Can GraspTune improve stable grasp execution across proposal generators and visual occlusion levels? \textbf{Q2:} Can GraspTune generalize to unseen object categories excluded from all training stages? \textbf{Q3:} Which representation, supervision, and policy-learning components produce GraspTune's gains? \textbf{Q4:} Can GraspTune transfer from simulation to real-robot grasp execution without real-world fine-tuning? Each experiment answers one of these questions using a matched evaluation protocol.

\subsection{Experimental Protocol}

We train the policy in simulation on balanced MuJoCo demonstrations that reach stable post-close contact. The dataset contains 20 object classes with 20 trajectories per class, yielding 400 trajectories and 68,672 state-action records. A class-stratified split assigns 18 trajectories per class to training and two to validation. The policy receives depth, tactile signals, execution state, and history; it excludes category labels. Stages I and II train for 50 epochs with \(\beta_c=0.05\) and \(\beta_p=0.10\) in Stage I. Stage III uses a 10-epoch warm start and 20,000 online interaction steps. All reported results use the final checkpoints at these fixed training budgets; we select no checkpoint based on test-benchmark performance.

The main benchmark uses a fixed 20-class object pool: apple, banana, baseball, cracker box, foam brick, golf ball, lemon, mustard bottle, orange, pear, peach, screwdriver, plum, potted meat can, racquetball, Rubik's cube, softball, strawberry, tennis ball, and tomato soup can. The objects vary in size, geometry, surface properties, and mass. Simulated occlusion experiments use floating, non-colliding occluders at \(0\%\), \(20\%\), and \(40\%\) nominal target-projection coverage. Their randomized projected positions create incomplete RGB-D proposal inputs while preserving object--gripper contact dynamics for execution evaluation.

The primary metric is \emph{stable grasp success}: the object must be lifted and held for 3 s after lift; transient lifts followed by drops are failures. We use normal-approximation intervals for differences in independent binomial proportions to construct aggregate 95\% confidence intervals.

We evaluate the final policy after GraspNet, Contact-GraspNet, AnyGrasp, and VGN. For each generator, \(20\) classes \(\times\) \(3\) occlusion levels \(\times\) \(100\) trials are run for both Raw and GraspTune, giving 6000 trials per controller and 48,000 executions. Both controllers start from the same admitted proposal and share camera, screening, collision, reachability, object-pool, lift, and stable-success settings. We report \(P(\text{stable hold}\mid\text{valid proposal admitted to execution})\), the stable-hold probability conditional on proposal admission, and the execution gain
\[
\begin{aligned}
\Delta p
&=P(\text{hold}\mid\text{admitted},\text{GraspTune})\\
&\quad-P(\text{hold}\mid\text{admitted},\text{Raw}).
\end{aligned}
\]

\begin{figure}[t]
    \centering
    \includegraphics[width=\columnwidth]{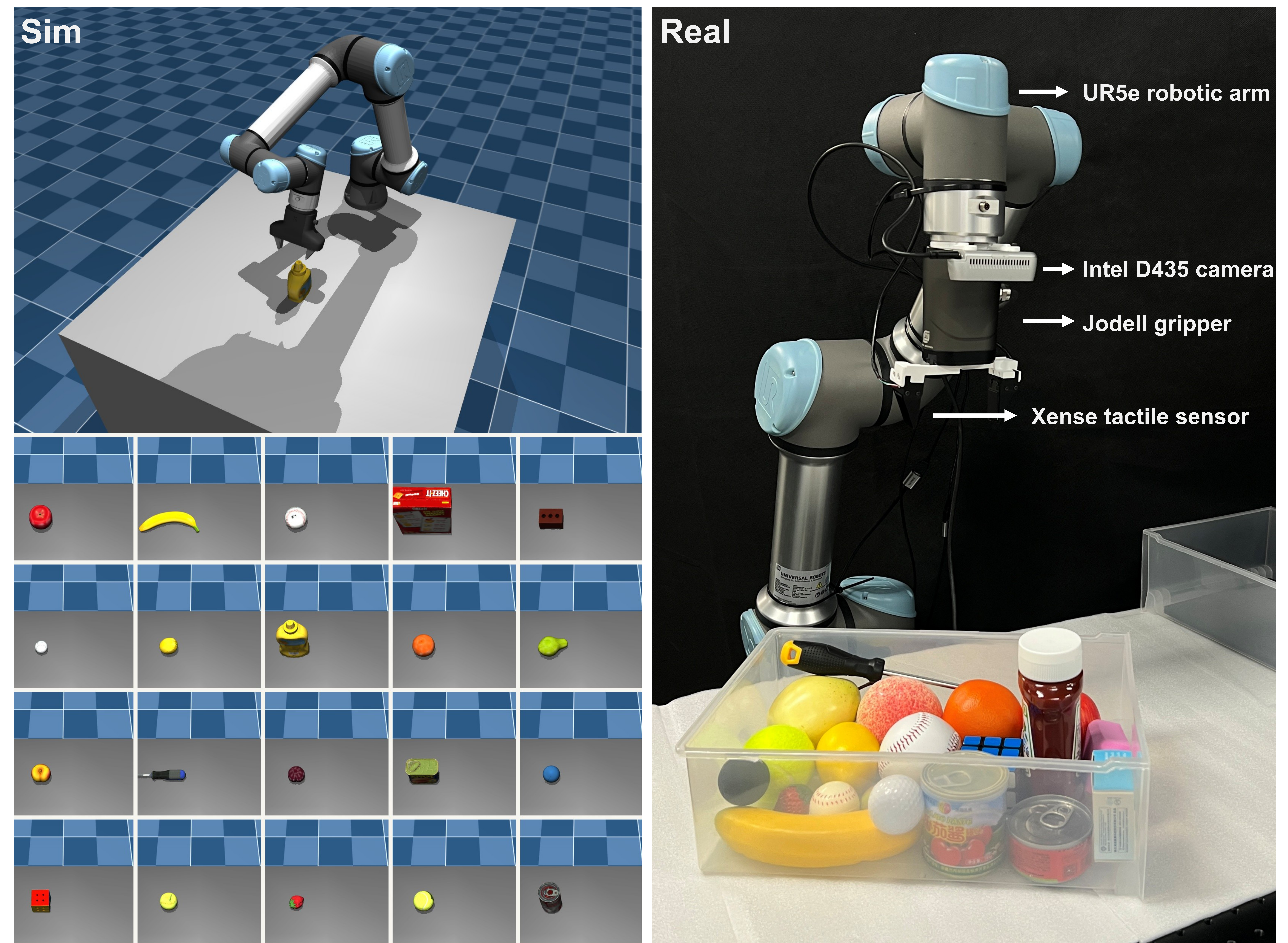}
    \caption{Simulation and real-robot evaluation setup, showing the MuJoCo scene, 20-object benchmark pool, and UR5e–Xense hardware platform.}
    \label{fig:experimental_setup}
\end{figure}

Fig.~\ref{fig:experimental_setup} links the controlled simulation benchmark and real-robot platform under the same execution objective: recover stable grasps from imperfect visual proposals using local tactile-driven feedback.

\begin{figure}[t]
    \centering
    \includegraphics[width=0.80\columnwidth]{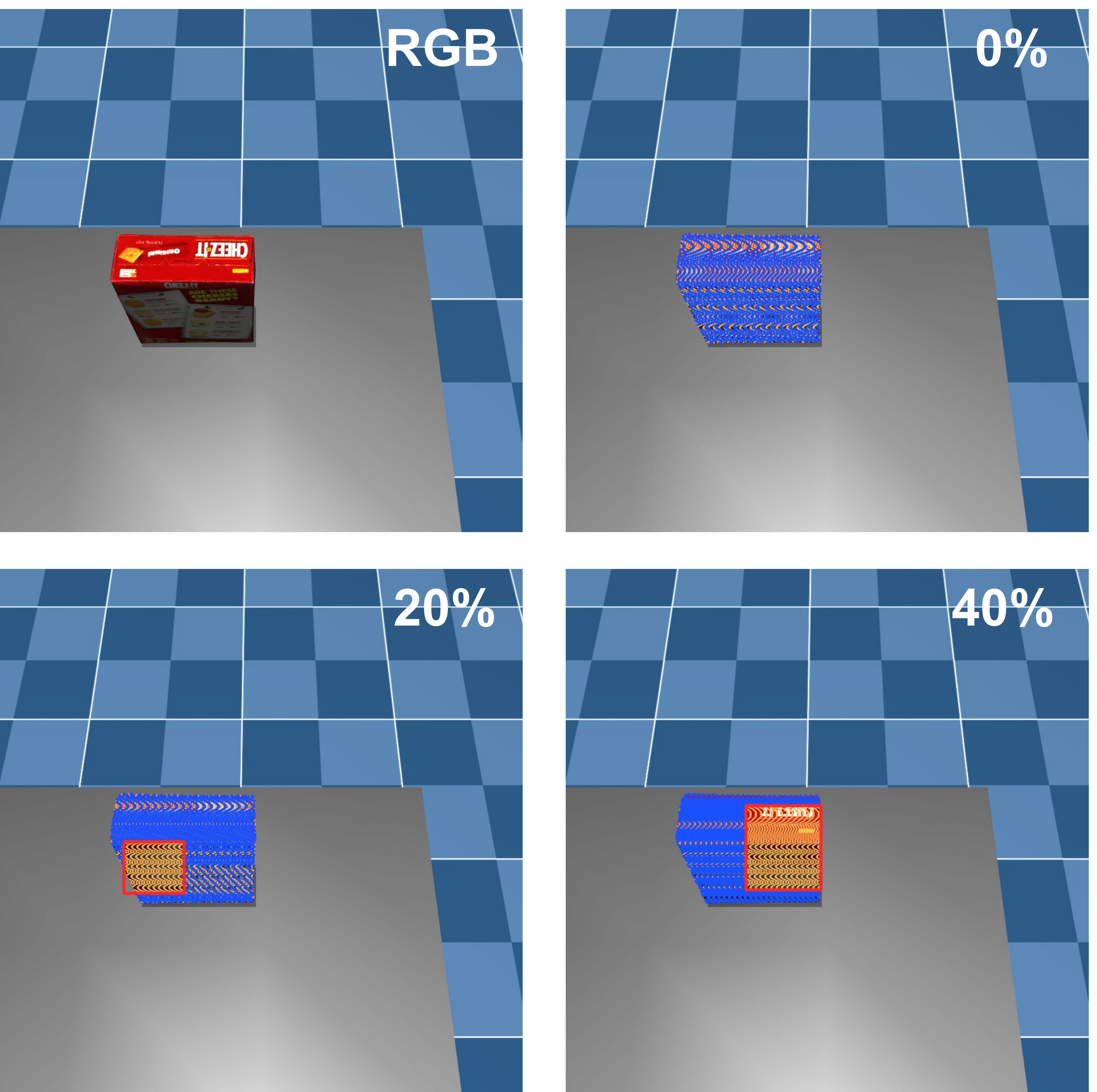}
    \caption{Physical floating occlusion and the resulting proposal inputs. Top Left: RGB observation. The remaining panels show proposal-input point clouds at nominal target-projection coverage levels of $0\%$, $20\%$, and $40\%$; blue marks observed target points, while orange marks the floating occluder.}
    \label{fig:occlusion_protocol}
\end{figure}

Fig.~\ref{fig:occlusion_protocol} defines the \(0\%\), \(20\%\), and \(40\%\) floating-occlusion levels used to vary proposal-stage visibility while preserving physical contact execution.

\subsection{Proposal Robustness}

GraspTune improves stable grasp execution for every evaluated proposal generator and at all three tested occlusion levels. Table~\ref{tab:proposal_main} shows gains from \(53.70\%\) to \(72.92\%\) for GraspNet, \(56.62\%\) to \(66.17\%\) for Contact-GraspNet, \(50.73\%\) to \(63.18\%\) for AnyGrasp, and \(50.10\%\) to \(70.80\%\) for VGN. The 95\% confidence intervals for the gains are strictly positive: \(+17.5\) to \(+20.9\) pp for GraspNet, \(+7.8\) to \(+11.3\) pp for Contact-GraspNet, \(+10.7\) to \(+14.2\) pp for AnyGrasp, and \(+19.0\) to \(+22.4\) pp for VGN. Because each proposal generator supplies its own admitted proposal distribution, this result establishes a common execution-layer benefit across point-cloud and volumetric proposal pipelines.

\begin{table}[t]
\caption{Stable grasp execution across proposal generators and visual occlusion levels}
\label{tab:proposal_main}
\centering
\scriptsize
\setlength{\tabcolsep}{2.2pt}
\begin{tabular}{@{}>{\raggedright\arraybackslash}p{0.36\columnwidth}cccc@{}}
\toprule
\textbf{Method} & \textbf{0\%} & \textbf{20\%} & \textbf{40\%} & \textbf{Overall} \\
\midrule
GraspNet raw & 66.10\% & 57.45\% & 37.55\% & 53.70\% \\
GraspNet + GraspTune & 83.95\% & 81.45\% & 53.35\% & \textbf{72.92\%} \\
\addlinespace[2pt]
Contact-GraspNet raw & 77.60\% & 52.60\% & 39.65\% & 56.62\% \\
Contact-GraspNet + GraspTune & 82.90\% & 70.55\% & 45.05\% & \textbf{66.17\%} \\
\addlinespace[2pt]
AnyGrasp raw & 65.80\% & 56.25\% & 30.15\% & 50.73\% \\
AnyGrasp + GraspTune & 86.10\% & 69.50\% & 33.95\% & \textbf{63.18\%} \\
\addlinespace[2pt]
VGN raw & 63.60\% & 43.95\% & 42.75\% & 50.10\% \\
VGN + GraspTune & 77.20\% & 68.70\% & 66.50\% & \textbf{70.80\%} \\
\bottomrule
\end{tabular}
\end{table}

With matched proposals, these gains isolate execution-stage refinement: GraspTune converts admitted visual grasps into stronger physical holds across diverse proposal generators.

\subsection{Held-out-category Transfer}

We partitioned the fixed 20-class object pool into four held-out category folds. For each fold, we trained the policy on 15 categories and evaluated it on the five unseen categories, using GraspNet proposals and the same three occlusion levels. We excluded held-out categories from Contact-Semantic Representation, Residual Policy Pretraining, and Policy Adaptation. The held-out sets were: A, lemon, plum, peach, softball, and Rubik's cube; B, foam brick, pear, golf ball, apple, and tomato soup can; C, racquetball, baseball, mustard bottle, screwdriver, and strawberry; and D, orange, tennis ball, potted meat can, cracker box, and banana. Each fold used the remaining 15 categories for training and 300 trials per method for testing.

GraspTune transfers its execution advantage to categories excluded from every training stage. Across 1200 held-out trials per method, Table~\ref{tab:heldout_category_generalization} shows that GraspTune raises raw GraspNet execution from \(54.58\%\) to \(70.33\%\), an absolute gain of \(15.75\) percentage points with a 95\% confidence interval of \(+11.9\) to \(+19.6\) pp. The aggregate gain is positive in all four category-disjoint folds, demonstrating that GraspTune learns transferable contact-correction behavior rather than category-specific execution templates.

\begin{table}[t]
\caption{Held-out-category transfer across four category-disjoint folds}
\label{tab:heldout_category_generalization}
\centering
\scriptsize
\begin{tabular}{@{}lcc@{}}
\toprule
\textbf{Execution} & \textbf{Stable success} & \textbf{Gain} \\
\midrule
GraspNet raw & 54.58\% & -- \\
GraspNet + GraspTune & \textbf{70.33\%} & \textbf{+15.75 pp} \\
\bottomrule
\end{tabular}
\end{table}

\begin{figure*}[t]
    \centering
    \includegraphics[width=1\textwidth]{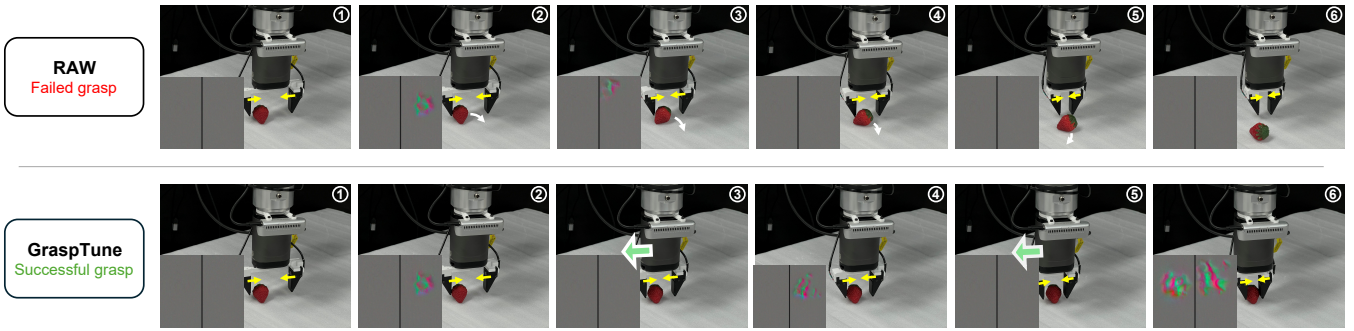}
    \caption{Same-proposal strawberry execution. Lower-left insets show the corresponding tactile Difference images. Yellow arrows indicate gripper closing, green arrows denote local residual TCP corrections, and white arrows indicate observed object motion.}
    \label{fig:qualitative_rollout}
\end{figure*}

\subsection{Component Ablations}

\begin{table}[t]
\caption{Component ablation on GraspNet proposals}
\label{tab:component_ablation}
\centering
\scriptsize
\setlength{\tabcolsep}{4pt}
\begin{tabular}{@{}>{\raggedright\arraybackslash}p{0.73\columnwidth}c@{}}
\toprule
\textbf{Variant} & \textbf{Overall success} \\
\midrule
Full model & \textbf{72.67\%} \\
\addlinespace[1pt]
\multicolumn{2}{@{}l}{\emph{Contact representation and tactile feedback}} \\
w/o contact-semantic representation & 61.42\% \\
w/o policy-visible tactile observations & 58.25\% \\
\addlinespace[1pt]
\multicolumn{2}{@{}l}{\emph{Contact-semantic supervision}} \\
w/o privileged-cue distillation & 57.00\% \\
w/o action-conditioned contact prediction & 51.75\% \\
\addlinespace[1pt]
\multicolumn{2}{@{}l}{\emph{Residual policy learning}} \\
w/ single-step BC actor \(\rightarrow\) PPO & 63.58\% \\
w/o policy adaptation & 60.33\% \\
w/o residual policy pretraining & 45.17\% \\
\addlinespace[1pt]
\bottomrule
\end{tabular}
\end{table}

All learned ablation variants use GraspNet proposals and the same balanced 20-class, three-occlusion protocol, evaluated on a fixed 1,200-trial proposal bank with a shared evaluation seed set. The full model reaches \(72.67\%\), compared with \(61.42\%\) without contact-semantic representation and \(58.25\%\) without policy-visible tactile observations. Removing privileged-cue distillation or action-conditioned contact prediction lowers success to \(57.00\%\) and \(51.75\%\), respectively. The action-conditioned ablation isolates the \(H=4\) action chunk in the Stage-I contact-query head while keeping contact-change, risk, readiness, diffusion-pretraining, distillation, and PPO settings fixed; the encoder architecture is unchanged, and Stage I and all downstream stages are retrained under the same budgets. The tactile-observation ablation zeroes tactile maps and masks tactile-derived contact variables from the policy-visible state and history while retaining depth, TCP pose, finger state, and proprioceptive execution variables.

The residual-policy-learning ablations share the same Stage-I encoder, demonstrations, residual action range, retained-stage training epochs, PPO reward, proposal bank, and 20,000-step PPO budget. Diffusion residual pretraining is decisive: removing it reduces PPO success from \(72.67\%\) to \(45.17\%\), a \(27.50\)-percentage-point drop under this matched protocol. A single-step BC actor followed by PPO reaches \(63.58\%\), and diffusion deployment without policy adaptation reaches \(60.33\%\). 

Table~\ref{tab:additional_validation} reports multi-seed results and the separate 6,000-trial comparison with the scripted tactile controller.

\begin{table}[t]
\caption{Multi-seed evaluation and scripted-controller comparison}
\label{tab:additional_validation}
\centering
\scriptsize
\setlength{\tabcolsep}{3pt}
\begin{tabular}{@{}>{\raggedright\arraybackslash}p{0.72\columnwidth}c@{}}
\toprule
\textbf{Setting} & \textbf{Stable success} \\
\midrule
Full GraspTune (3 seeds) & \(71.97 \pm 0.64\%\) \\
\addlinespace[1pt]
Scripted tactile controller (6,000 trials) & \(3657/6000 = 60.95\%\) \\
GraspTune (6,000 trials) & \textbf{\(4375/6000 = 72.92\%\)} \\
\bottomrule
\end{tabular}
\end{table}

\subsection{Real-robot Transfer}

We evaluate GraspTune on a UR5e arm with a Jodell parallel gripper, fingertip-mounted Xense visuo-tactile sensors, and an eye-in-hand Intel RealSense D435 camera. The real-robot protocol compares raw GraspNet and GraspNet+GraspTune under the same hand-eye calibration, mapped 6-DoF execution, synchronized Xense logging, and tactile preprocessing.

The simulation-trained policy transfers directly to physical execution, raising stable grasp success from \(426/600=71.0\%\) to \(506/600=84.3\%\), a \(13.3\)-percentage-point gain with a 95\% confidence interval of \(+8.7\) to \(+18.0\) pp. The evaluation uses the same 20 semantic object categories as in the simulation, with physical instances spanning different weights, materials, surface frictions, local geometries, and tactile contact distributions. Under a clean \(0\%\)-occlusion tabletop setting, each category receives 30 trials for raw GraspNet and GraspNet+GraspTune. Table~\ref{tab:real_robot_occ0} gives the aggregate direct-transfer gain, and Table~\ref{tab:real_per_class} shows improvement or maintained success across all 20 categories.

\begin{table}[t]
\caption{Real-robot direct transfer under \(0\%\) occlusion}
\label{tab:real_robot_occ0}
\centering
\scriptsize
\begin{tabular}{@{}lccc@{}}
\toprule
\textbf{Execution} & \textbf{Success trials} & \textbf{Success rate} & \textbf{Gain} \\
\midrule
Raw GraspNet & \(426/600\) & \(71.0\%\) & -- \\
GraspNet + GraspTune & \(\mathbf{506/600}\) & \(\mathbf{84.3\%}\) & \(\mathbf{+13.3}\) pp \\
\bottomrule
\end{tabular}
\end{table}

\begin{table}[t]
\caption{Per-category real-robot stable-grasp success (30 trials per category)}
\label{tab:real_per_class}
\centering
\scriptsize
\setlength{\tabcolsep}{2.4pt}
\begin{tabular}{@{}lcc@{\hspace{0.7em}}lcc@{}}
\toprule
\textbf{Category} & \textbf{Raw} & \textbf{GraspTune} & \textbf{Category} & \textbf{Raw} & \textbf{GraspTune} \\
\midrule
apple & 24 & 28 & peach & 26 & 28 \\
banana & 22 & 25 & screwdriver & 20 & 26 \\
baseball & 26 & 28 & plum & 6 & 15 \\
cracker box & 27 & 30 & potted meat & 23 & 29 \\
foam brick & 30 & 30 & racquetball & 12 & 18 \\
golf ball & 27 & 28 & rubiks cube & 29 & 30 \\
lemon & 23 & 27 & softball & 25 & 30 \\
mustard bottle & 9 & 16 & strawberry & 7 & 13 \\
orange & 24 & 26 & tennis ball & 23 & 27 \\
pear & 19 & 24 & tomato soup & 24 & 28 \\
\bottomrule
\end{tabular}
\end{table}

Fig.~\ref{fig:qualitative_rollout} follows a strawberry from the same off-center nominal proposal. The lower-left inset in each panel shows the corresponding tactile Difference image. In Raw execution, direct closure produces asymmetric contact, and the strawberry rolls out of the closing region before lift. In contrast, GraspTune uses the emerging contact asymmetry as feedback for local correction. After unilateral contact emerges, GraspTune translates the TCP away from the stronger-contact side, relieving that contact while creating room to establish support on the opposing finger. The resulting bilateral contact stabilizes the strawberry before final closure and enables a successful lift.

\section{Conclusions and Limitations}

GraspTune treats grasping as tactile-driven physical execution rather than a one-shot visual decision. Contact-Semantic Representation learns control-facing tactile evidence, Residual Policy Pretraining turns it into a residual behavior model, and Policy Adaptation converts it into a closed-loop controller. Across more than 60,000 simulated executions, gains of \(+19.22\), \(+9.55\), \(+12.45\), and \(+20.70\) percentage points after GraspNet, Contact-GraspNet, AnyGrasp, and VGN establish execution-layer effectiveness across proposal generators. The held-out-category gain from \(54.58\%\) to \(70.33\%\) shows that the learned contact corrections transfer beyond trained categories. Without real-world policy fine-tuning, the real-robot gain from \(71.0\%\) to \(84.3\%\) validates direct transfer to physical grasp execution. Robust grasping needs strong visual proposals and tactile-driven execution that converts them into stable physical grasps.

\textit{Limitations and future work.} The direct-transfer interface is calibrated for a \(100\) Hz simulated loop and an approximately \(10\) Hz Xense--policy--UR5e--gripper loop, with Xense Difference responses mapped to the encoder's indentation-equivalent scale. GraspTune controls residual translation and yaw, while a contact-dependent schedule governs closure. Future work will extend the interface across control rates, communication latencies, tactile response models, and hardware stacks, and will jointly optimize TCP motion and closure for the aperture, contact-geometry, and timing couplings of parallel-jaw and forward-reaching grippers.

\bibliographystyle{IEEEtran}
\bibliography{reflex_refs}

@inproceedings{fang2020graspnet,
  author    = {Hao-Shu Fang and Chenxi Wang and Minghao Gou and Cewu Lu},
  title     = {{GraspNet-1Billion}: A Large-Scale Benchmark for General Object Grasping},
  booktitle = {Proceedings of the IEEE/CVF Conference on Computer Vision and Pattern Recognition (CVPR)},
  pages     = {11441--11450},
  year      = {2020},
  doi       = {10.1109/CVPR42600.2020.01146}
}

@article{fang2023anygrasp,
  author  = {Hao-Shu Fang and Chenxi Wang and Hongjie Fang and Minghao Gou and Jirong Liu and Hengxu Yan and Wenhai Liu and Yichen Xie and Cewu Lu},
  title   = {{AnyGrasp}: Robust and Efficient Grasp Perception in Spatial and Temporal Domains},
  journal = {IEEE Transactions on Robotics},
  volume  = {39},
  number  = {5},
  pages   = {3929--3945},
  year    = {2023},
  doi     = {10.1109/TRO.2023.3281153}
}

@inproceedings{sundermeyer2021contactgraspnet,
  author    = {Martin Sundermeyer and Arsalan Mousavian and Rudolph Triebel and Dieter Fox},
  title     = {{Contact-GraspNet}: Efficient 6-{DoF} Grasp Generation in Cluttered Scenes},
  booktitle = {Proceedings of the IEEE International Conference on Robotics and Automation (ICRA)},
  pages     = {13438--13444},
  year      = {2021},
  doi       = {10.1109/ICRA48506.2021.9561877}
}

@inproceedings{breyer2021vgn,
  author    = {Michel Breyer and Jen Jen Chung and Lionel Ott and Roland Siegwart and Juan Nieto},
  title     = {Volumetric Grasping Network: Real-Time 6-{DOF} Grasp Detection in Clutter},
  booktitle = {Proceedings of the 2020 Conference on Robot Learning},
  series    = {Proceedings of Machine Learning Research},
  volume    = {155},
  pages     = {1602--1611},
  publisher = {PMLR},
  year      = {2021},
  url       = {https://proceedings.mlr.press/v155/breyer21a.html}
}

@inproceedings{jiang2021giga,
  author    = {Zhenyu Jiang and Yifeng Zhu and Maxwell Svetlik and Kuan Fang and Yuke Zhu},
  title     = {Synergies Between Affordance and Geometry: 6-{DoF} Grasp Detection via Implicit Representations},
  booktitle = {Proceedings of Robotics: Science and Systems},
  year      = {2021},
  doi       = {10.15607/RSS.2021.XVII.024}
}

@inproceedings{dai2023graspnerf,
  author    = {Qiyu Dai and Yan Zhu and Yiran Geng and Ciyu Ruan and Jiazhao Zhang and He Wang},
  title     = {{GraspNeRF}: Multiview-Based 6-{DoF} Grasp Detection for Transparent and Specular Objects Using Generalizable {NeRF}},
  booktitle = {Proceedings of the IEEE International Conference on Robotics and Automation (ICRA)},
  pages     = {1757--1763},
  year      = {2023},
  doi       = {10.1109/ICRA48891.2023.10160842}
}

@inproceedings{chi2023diffusion,
  author    = {Cheng Chi and Siyuan Feng and Yilun Du and Zhenjia Xu and Eric Cousineau and Benjamin C. M. Burchfiel and Shuran Song},
  title     = {Diffusion Policy: Visuomotor Policy Learning via Action Diffusion},
  booktitle = {Proceedings of Robotics: Science and Systems},
  year      = {2023},
  doi       = {10.15607/RSS.2023.XIX.026}
}

@article{tenpas2017gpd,
  author  = {Andreas ten Pas and Marcus Gualtieri and Kate Saenko and Robert Platt},
  title   = {Grasp Pose Detection in Point Clouds},
  journal = {The International Journal of Robotics Research},
  volume  = {36},
  number  = {13--14},
  pages   = {1455--1473},
  year    = {2017},
  doi     = {10.1177/0278364917735594}
}

@inproceedings{mahler2017dexnet,
  author    = {Jeffrey Mahler and Jacky Liang and Sherdil Niyaz and Michael Laskey and Richard Doan and Xinyu Liu and Juan Aparicio Ojea and Ken Goldberg},
  title     = {{Dex-Net} 2.0: Deep Learning to Plan Robust Grasps with Synthetic Point Clouds and Analytic Grasp Metrics},
  booktitle = {Proceedings of Robotics: Science and Systems},
  year      = {2017},
  doi       = {10.15607/RSS.2017.XIII.058}
}

@inproceedings{liang2019pointnetgpd,
  author    = {Hongzhuo Liang and Xiaojian Ma and Shuang Li and Michael G{\"o}rner and Song Tang and Bin Fang and Fuchun Sun and Jianwei Zhang},
  title     = {{PointNetGPD}: Detecting Grasp Configurations from Point Sets},
  booktitle = {Proceedings of the IEEE International Conference on Robotics and Automation (ICRA)},
  pages     = {3629--3635},
  year      = {2019},
  doi       = {10.1109/ICRA.2019.8794435}
}

@article{mahler2019ambidextrous,
  author   = {Jeffrey Mahler and Matthew Matl and Vishal Satish and Michael Danielczuk and Bill DeRose and Stephen McKinley and Ken Goldberg},
  title    = {Learning Ambidextrous Robot Grasping Policies},
  journal  = {Science Robotics},
  volume   = {4},
  number   = {26},
  pages    = {eaau4984},
  year     = {2019},
  doi      = {10.1126/scirobotics.aau4984}
}

@inproceedings{mousavian20196dofgraspnet,
  author    = {Arsalan Mousavian and Clemens Eppner and Dieter Fox},
  title     = {6-{DOF} {GraspNet}: Variational Grasp Generation for Object Manipulation},
  booktitle = {Proceedings of the IEEE/CVF International Conference on Computer Vision (ICCV)},
  pages     = {2901--2910},
  year      = {2019},
  doi       = {10.1109/ICCV.2019.00299}
}

@inproceedings{qin2020s4g,
  author    = {Yuzhe Qin and Rui Chen and Hao Zhu and Meng Song and Jing Xu and Hao Su},
  title     = {{S4G}: Amodal Single-View Single-Shot {SE(3)} Grasp Detection in Cluttered Scenes},
  booktitle = {Proceedings of the Conference on Robot Learning},
  series    = {Proceedings of Machine Learning Research},
  volume    = {100},
  pages     = {53--65},
  publisher = {PMLR},
  year      = {2020},
  url       = {https://proceedings.mlr.press/v100/qin20a.html}
}

@article{levine2018handeye,
  author  = {Sergey Levine and Peter Pastor and Alex Krizhevsky and Julian Ibarz and Deirdre Quillen},
  title   = {Learning Hand-Eye Coordination for Robotic Grasping with Deep Learning and Large-Scale Data Collection},
  journal = {The International Journal of Robotics Research},
  volume  = {37},
  number  = {4--5},
  pages   = {421--436},
  year    = {2018},
  doi     = {10.1177/0278364917710318}
}

@inproceedings{kalashnikov2018qtopt,
  author    = {Dmitry Kalashnikov and Alex Irpan and Peter Pastor and Julian Ibarz and Alexander Herzog and Eric Jang and Deirdre Quillen and Ethan Holly and Mrinal Kalakrishnan and Vincent Vanhoucke and Sergey Levine},
  title     = {Scalable Deep Reinforcement Learning for Vision-Based Robotic Manipulation},
  booktitle = {Proceedings of the 2nd Conference on Robot Learning},
  series    = {Proceedings of Machine Learning Research},
  volume    = {87},
  pages     = {651--673},
  publisher = {PMLR},
  year      = {2018},
  url       = {https://proceedings.mlr.press/v87/kalashnikov18a.html}
}

@inproceedings{lee2019making,
  author    = {Michelle A. Lee and Yuke Zhu and Krishnan Srinivasan and Parth Shah and Silvio Savarese and Li Fei-Fei and Animesh Garg and Jeannette Bohg},
  title     = {Making Sense of Vision and Touch: Self-Supervised Learning of Multimodal Representations for Contact-Rich Tasks},
  booktitle = {Proceedings of the IEEE International Conference on Robotics and Automation (ICRA)},
  pages     = {8943--8950},
  year      = {2019},
  doi       = {10.1109/ICRA.2019.8793485}
}

@inproceedings{tian2019manipulation,
  author    = {Stephen Tian and Frederik Ebert and Dinesh Jayaraman and Mayur Mudigonda and Chelsea Finn and Roberto Calandra and Sergey Levine},
  title     = {Manipulation by Feel: Touch-Based Control with Deep Predictive Models},
  booktitle = {Proceedings of the IEEE International Conference on Robotics and Automation (ICRA)},
  pages     = {818--824},
  year      = {2019},
  doi       = {10.1109/ICRA.2019.8794219}
}

@inproceedings{hogan2020tactile,
  author    = {Francois R. Hogan and Jose Ballester and Siyuan Dong and Alberto Rodriguez},
  title     = {Tactile Dexterity: Manipulation Primitives with Tactile Feedback},
  booktitle = {Proceedings of the IEEE International Conference on Robotics and Automation (ICRA)},
  pages     = {8863--8869},
  year      = {2020},
  doi       = {10.1109/ICRA40945.2020.9196976}
}

@article{schulman2017ppo,
  author        = {John Schulman and Filip Wolski and Prafulla Dhariwal and Alec Radford and Oleg Klimov},
  title         = {Proximal Policy Optimization Algorithms},
  journal       = {arXiv preprint arXiv:1707.06347},
  year          = {2017},
  eprint        = {1707.06347},
  archivePrefix = {arXiv},
  primaryClass  = {cs.LG},
  doi           = {10.48550/arXiv.1707.06347}
}

@inproceedings{morrison2018ggcnn,
  author    = {Douglas Morrison and Peter Corke and J{\"u}rgen Leitner},
  title     = {Closing the Loop for Robotic Grasping: A Real-Time, Generative Grasp Synthesis Approach},
  booktitle = {Proceedings of Robotics: Science and Systems},
  year      = {2018},
  doi       = {10.15607/RSS.2018.XIV.021}
}

@article{haviland2020finalphase,
  author        = {Jesse Haviland and Feras Dayoub and Peter Corke},
  title         = {Control of the Final-Phase of Closed-Loop Visual Grasping Using Image-Based Visual Servoing},
  journal       = {arXiv preprint arXiv:2001.05650},
  year          = {2020},
  eprint        = {2001.05650},
  archivePrefix = {arXiv},
  primaryClass  = {cs.RO},
  doi           = {10.48550/arXiv.2001.05650}
}

@inproceedings{li2013tactileservoing,
  author    = {Qiang Li and Carsten Sch{\"u}rmann and Robert Haschke and Helge Ritter},
  title     = {A Control Framework for Tactile Servoing},
  booktitle = {Proceedings of Robotics: Science and Systems},
  year      = {2013},
  doi       = {10.15607/RSS.2013.IX.045}
}

@article{calandra2018more,
  author  = {Roberto Calandra and Andrew Owens and Dinesh Jayaraman and Justin Lin and Wenzhen Yuan and Jitendra Malik and Edward H. Adelson and Sergey Levine},
  title   = {More Than a Feeling: Learning to Grasp and Regrasp Using Vision and Touch},
  journal = {IEEE Robotics and Automation Letters},
  volume  = {3},
  number  = {4},
  pages   = {3300--3307},
  year    = {2018},
  doi     = {10.1109/LRA.2018.2852779}
}

@article{wang2025tacrefinenet,
  author        = {Shuaijun Wang and Haoran Zhou and Diyun Xiang and Yangwei You},
  title         = {{TacRefineNet}: Goal-Conditioned Tactile Grasp Refinement for Edge-Prominent Objects},
  journal       = {arXiv preprint arXiv:2509.25746},
  year          = {2025},
  eprint        = {2509.25746},
  archivePrefix = {arXiv},
  primaryClass  = {cs.RO},
  doi           = {10.48550/arXiv.2509.25746}
}

@article{li2026goto,
  author  = {Juntao Li and Qihang Fan and Xingke Xia and Daqiang Guo},
  title   = {Glance Once-Touch Once ({GOTO}): Physics-Informed Visuo-Tactile Human-Like Grasping for Embodied Robotic Manufacturing},
  journal = {Robotics and Computer-Integrated Manufacturing},
  volume  = {103},
  pages   = {103357},
  year    = {2027},
  doi     = {10.1016/j.rcim.2026.103357}
}

@article{wang2025d3grasp,
  author        = {Keyu Wang and Bingcong Lu and Zhengxue Cheng and Hengdi Zhang and Li Song},
  title         = {{D3Grasp}: Diverse and Deformable Dexterous Grasping for General Objects},
  journal       = {arXiv preprint arXiv:2509.19892},
  year          = {2025},
  eprint        = {2509.19892},
  archivePrefix = {arXiv},
  primaryClass  = {cs.RO},
  doi           = {10.48550/arXiv.2509.19892}
}

@article{zhang2026touchguide,
  author        = {Zhemeng Zhang and Jiahua Ma and Xincheng Yang and Xin Wen and Yuzhi Zhang and Boyan Li and Yiran Qin and Jin Liu and Can Zhao and Li Kang and Haoqin Hong and Zhenfei Yin and Philip Torr and Hao Su and Ruimao Zhang and Daolin Ma},
  title         = {{TouchGuide}: Inference-Time Steering of Visuomotor Policies via Touch Guidance},
  journal       = {arXiv preprint arXiv:2601.20239},
  year          = {2026},
  eprint        = {2601.20239},
  archivePrefix = {arXiv},
  primaryClass  = {cs.RO},
  doi           = {10.48550/arXiv.2601.20239}
}

@inproceedings{xue2025rdp,
  author    = {Han Xue and Jieji Ren and Wendi Chen and Gu Zhang and Fang Yuan and Guoying Gu and Huazhe Xu and Cewu Lu},
  title     = {Reactive Diffusion Policy: Slow-Fast Visual-Tactile Policy Learning for Contact-Rich Manipulation},
  booktitle = {Proceedings of Robotics: Science and Systems},
  year      = {2025},
  doi       = {10.15607/RSS.2025.XXI.052}
}

@article{ma2026taccorl,
  author        = {Siyu Ma and Yuqi Liang and Chang Yu and Yunuo Chen and Hao Su and Yixin Zhu and Yin Yang and Chenfanfu Jiang},
  title         = {{TacCoRL}: Integrating Tactile Feedback into {VLA} via Simulation},
  journal       = {arXiv preprint arXiv:2606.11743},
  year          = {2026},
  eprint        = {2606.11743},
  archivePrefix = {arXiv},
  primaryClass  = {cs.RO},
  doi           = {10.48550/arXiv.2606.11743}
}

@article{yu2026omnitactune,
  author        = {Kelin Yu and Haode Zhang and Harish Ravichandar and Yunhai Han and Ruohan Gao},
  title         = {{OmniTacTune}: Policy-Agnostic Real-World {RL} for Tactile Residual Adaptation of Visual Policies},
  journal       = {arXiv preprint arXiv:2607.03723},
  year          = {2026},
  eprint        = {2607.03723},
  archivePrefix = {arXiv},
  primaryClass  = {cs.RO},
  doi           = {10.48550/arXiv.2607.03723}
}

\end{document}